\documentclass[11pt]{article}

\usepackage[final]{acl}
\usepackage{algorithm}
\usepackage{algpseudocode}
\usepackage{amsmath}
\usepackage{amsfonts}
\usepackage{times}
\usepackage{latexsym}
\usepackage[T1]{fontenc}
\usepackage[utf8]{inputenc}
\usepackage{multirow}
\usepackage{microtype}

\usepackage{inconsolata}

\usepackage{graphicx}
\usepackage{adjustbox}
\usepackage{booktabs}
\usepackage{placeins}
\title{Evolutionary Strategies lead to Catastrophic Forgetting in LLMs}

\author{\textbf{Immanuel Abdi\thanks{Equal contribution.}, Akshat Gupta\footnotemark[1]}\\ \textbf{Micah Mok, Alexander Lu, Nicholas Lee, Gopala Anumanchipalli}\\
  UC Berkeley \\
  \texttt{\{immanuelazn, akshat.gupta\}@berkeley.edu}
}

\begin{document}
\maketitle


\begin{abstract}
One of the biggest missing capabilities in current AI systems is the ability to learn continuously after deployment. Implementing such continually learning systems have several challenges, one of which is the large memory requirement of gradient-based algorithms that are used to train state-of-the-art LLMs. Evolutionary Strategies (ES) have recently re-emerged as a gradient-free alternative to traditional learning algorithms and have shown encouraging performance on specific tasks in LLMs. In this paper, we perform a comprehensive analysis of ES and specifically evaluate its forgetting curves when training for an increasing number of update steps. We first find that ES is able to reach performance numbers close to GRPO for math and reasoning tasks with a comparable compute budget. However, and most importantly for continual learning, \textbf{the performance gains in ES is accompanied by significant forgetting of prior abilities, limiting its applicability for training models online}. We also explore the reason behind this behavior and show that the updates made using ES are much less sparse and have orders of magnitude larger $\ell_2$ norm compared to corresponding GRPO updates, explaining the contrasting forgetting curves between the two algorithms. With this study, we aim to highlight the issue of forgetting in gradient-free algorithms like ES and hope to inspire future work to mitigate these issues. 
\end{abstract}

\section{Introduction}

Despite rapid advances in AI with transformer-based LLMs \citep{vaswani2017attention, brown2020language, deepseekai2024deepseekllm}, most state-of-the-art systems remain static after training and lack the ability to learn continually during deployment. In many real-world settings, models need to adapt to new tasks, user preferences, or data distributions to perform optimally. While modern chatbots like ChatGPT do this by taking notes in the form of user memory \cite{openai_memory_2024} and use in-context learning \cite{brown2020language} to incorporate this information, we currently lack solutions that can achieve this by modifying the model weights during deployment. One of the reasons that makes this challenging is that current post-training and adaptation methods for LLMs are exclusively gradient-based, including approaches such as SFT \cite{wei2022finetunedlanguagemodelszeroshot}, RLHF \citep{ouyangTrainingLanguageModels2022}, DPO \cite{rafailov2024directpreferenceoptimizationlanguage}, and GRPO \citep{shao2024deepseekmathpushinglimitsmathematical}. While effective, these methods require storing gradients, optimizer states, or intermediate activations, causing substantial memory overhead. 

Evolutionary Strategies (ES) \cite{qiuEvolutionStrategiesScale2025, korotyshova2025essaevolutionarystrategiesscalable} have recently re-emerged as a gradient-free alternative for optimizing LLMs. By estimating updates through population-level perturbations rather than backpropagation, ES avoids explicit gradient storage and can significantly reduce memory requirements during deployment. \citet{qiuEvolutionStrategiesScale2025} have shown that ES achieves comparable performance to GRPO on the Countdown task \cite{panJiayiPanTinyZero2026}, presenting ES as a viable candidate for continual learning in LLMs. However, a more comprehensive analysis on task generalization was missing in their work. More importantly from the perspective of continual learning, \citet{qiuEvolutionStrategiesScale2025} do not evaluate the extent to which ES preserves existing capabilities while learning new tasks. 

In this work, we present a comprehensive empirical analysis of ES for fine-tuning LLMs, with a focus on continual learning and forgetting. We compare ES against GRPO on multiple math and reasoning benchmarks and evaluate forgetting curves over many update steps. Our results confirm that ES is able to reach performance levels comparable to GRPO on a large suite of tasks; however, contrary to results reported in \citet{qiuEvolutionStrategiesScale2025}, we find that GRPO still dominates ES marginally on almost all tasks. Additionally, we show that training LLMs using ES leads to significant model degradation and forgetting of existing abilities when compared to GRPO. To better understand this behavior, we analyze the structure of parameter updates produced by ES and compare them to those obtained using GRPO. We find that ES updates are significantly less sparse and exhibit much larger $\ell_2$ norms, leading to more global parameter changes that interfere with previously learned capabilities.

Our findings highlight that although ES presents a tempting memory-efficient and gradient-free alternative to inference-time model adaptation, it is also accompanied by ``catastrophic'' forgetting \cite{kirkpatrick2017overcoming, gupta2024model} of prior abilities of the model. We hope these results can inspire future advancements in gradient-free algorithms with continual learning and catastrophic forgetting at the forefront of thought. We also release our codebase\footnote{Our codeabase can be found here - \url{https://github.com/akshat57/es-catastrophic}} and models\footnote{Our models can be found here -  \url{https://huggingface.co/collections/immanuelabdi/es-at-scale-lead-to-catastrophic-forgetting}} for reference.

To summarize, our work makes the following contributions:

\begin{enumerate}
    \item We show that ES is able to reach comparable performance to GRPO on several math and reasoning benchmarks with similar number of update steps.
    \item We show that training models using ES causes significant model degradation when compared to GRPO, leading to catastrophic fortgetting of prior abilities.
    \item Finally, we also explore the reason behind catastrophic forgetting in ES and show that this happens because model updates using ES are much less sparse when compared to GRPO with significantly larger $\ell_2$ norms. 
\end{enumerate}

\section{Related Work}
Evolution Strategies are a class of algorithms that search for solutions to first-order optimization problems by randomly modifying population members to find better performing members \citep{10.1007/978-3-642-83814-9_6, schwefel1977numerische, Beyer1995b}. Although implementations such as CMA-ES \cite{hansen2001completelyderandomizedselfadaptationinevolutionstrategies} and natural ES \cite{wierstra2011naturalevolutionstrategies, sun2012efficientnaturalevolutionstrategies} demonstrated success, initial implementations remained in the million-parameter scale \citep{such2018deepneuroevolutiongeneticalgorithms, risi2019deepneuroevolutionrecurrentdiscrete, zhang2017relationshipopenaievolutionstrategy}. However, recent updates have brought ES up to scale and in competition with GPRO, leveraging how it is highly parallelizable, \cite{salimansEvolutionStrategiesScalable2017}, memory efficient \cite{malladi2024finetuninglanguagemodelsjust, korotyshova2025essaevolutionarystrategiesscalable}, faster \cite{sarkar2025evolutionstrategieshyperscale}, robust to sparse reward horizons \cite{salimansEvolutionStrategiesScalable2017}, and can be modified with LoRA adaptions \cite{jin2024derivativefreeoptimizationlowrankadaptation, korotyshova2025essaevolutionarystrategiesscalable, sarkar2025evolutionstrategieshyperscale}. 
\citet{qiuEvolutionStrategiesScale2025} recently published a novel implementation of ES and showed that it outperforms GRPO. However, their study lacked a thorough analysis of model degradation during continued training. Additionally, a bulk of their study was focused on a single dataset. We extend their analysis to multiple datasets, evaluate model degradation during fine-tuning and also study the difference in weight updates in ES when compared to GRPO.

\section{Experiments}
\subsection{ES vs GRPO Comparison} \label{subsec:experiment-datasets}
We use the ES implementation of \citet{qiuEvolutionStrategiesScale2025} and compare it with the GRPO \cite{shao2024deepseekmathpushinglimitsmathematical} implementation from the VERL libary \citep{Sheng_2025}. An algorithmic analogy between the ES and GRPO algorithms can be found in \ref{appendix:A.1} while implementations details can be found in \ref{appendix:A.4}. We extend the analysis of ES and GRPO to three math and reasoning tasks -- GSM8K \citep{cobbe2021trainingverifierssolvemath}, MATH \citep{hendrycks2021measuringmathematicalproblemsolving} and OlympiadBench \citep{he2024olympiadbenchchallengingbenchmarkpromoting}, in addition to the Countdown dataset which was extensively studied in prior work \cite{qiuEvolutionStrategiesScale2025}. We perform this study for two models: Qwen2.5-1.5B-Instruct \citep{qwenQwen25TechnicalReport2024} and Llama-3.2-1B-Instruct \citep{grattafiori2024llama3herdmodels}. Following the experimental conditions of \citet{qiuEvolutionStrategiesScale2025}, we train our models on 200 examples from each dataset with identical batch size and number of rollouts. 

The results for comparison between ES and GRPO for fine-tuning LLMs can be found in Table \ref{tab:dataset-accuracy}. We see that for both models, ES is within 3-4 percentage points of GRPO in terms of task performance. These results are in contrast to prior work by \citet{qiuEvolutionStrategiesScale2025}, who claim that ES significantly outperforms GRPO on the Countdown task. In our experiments, we see that although ES performance numbers are close to GRPO, GRPO still outperforms ES for all but the GSM8K dataset with Llama-3.2-1B model. Therefore, we find different relative performance trends than those reported in prior work, which may stem from differences in GRPO implementations, hyperparameter choices, or evaluation protocols. We release our codebase and open source our trained models for reference.

\begin{table}[t]
\centering
\scalebox{0.8}{
\begin{tabular}{llcc}
\toprule
Model & Task & ES & GRPO \\
\midrule
\multirow{4}{*}{\shortstack[l]{Qwen-2.5-1.5B\\ (Instruct)}}
 & Countdown       & 53.0 & \textbf{56.4} \\
 & GSM8K         & 77.4 & \textbf{80.4} \\
 & MATH            & 59.1 & \textbf{63.2} \\
 & OlympiadBench   & 15.2 & \textbf{18.2} \\
\midrule
\multirow{4}{*}{\shortstack[l]{Llama-3.2-1B\\ (Instruct)}}
 & Countdown       & 15.2 & \textbf{37.6} \\
 & GSM8K         & \textbf{55.2} & 53.8 \\
 & MATH            & 32.2 & \textbf{35.6} \\
 & OlympiadBench   & 5.6  & \textbf{5.9} \\
\bottomrule
\end{tabular}
}
\caption{Peak validation accuracy (\%) across tasks and models using previously found optimal hyperparameters. The same hyperparameters were used across ES runs and across GRPO runs.}
\label{tab:dataset-accuracy}
\end{table}

The fact that the performance numbers of ES are comparable to a state-of-the-art post-training algorithm like GRPO is very encouraging and establishes ES as a potential gradient-free alternative to training LLMs. We also see that for all tasks except Countdown, ES is able to reach peak performance in similar number of update steps, which is shown in Figure \ref{fig:es-grpo-comparison}. This makes the compute requirements of ES also comparable to GRPO.

\subsection{ES and Catastrophic Forgetting} \label{subsec:exp-catastrophic-forgetting}

While Section \ref{subsec:experiment-datasets} shows that ES performs competitively with GRPO on various downstream tasks, a defining factor in the viability of using a fine-tuning algorithm for continual learning is its relationship with catastrophic forgetting. We utilized Qwen2.5-1.5B-Instruct trained on the Countdown dataset with GRPO and ES to evaluate catastrophic forgetting.  HellaSwag \citep{zellers2019hellaswagmachinereallyfinish} was used to evaluate LLMs on their prior capabilities. In an ideal scenario, performance on previous tasks should be preserved as new capability is gained. We thus evaluate task performance across each checkpoint of our trained models.

\begin{figure}[t]
\centering
\includegraphics[width=0.9\linewidth]{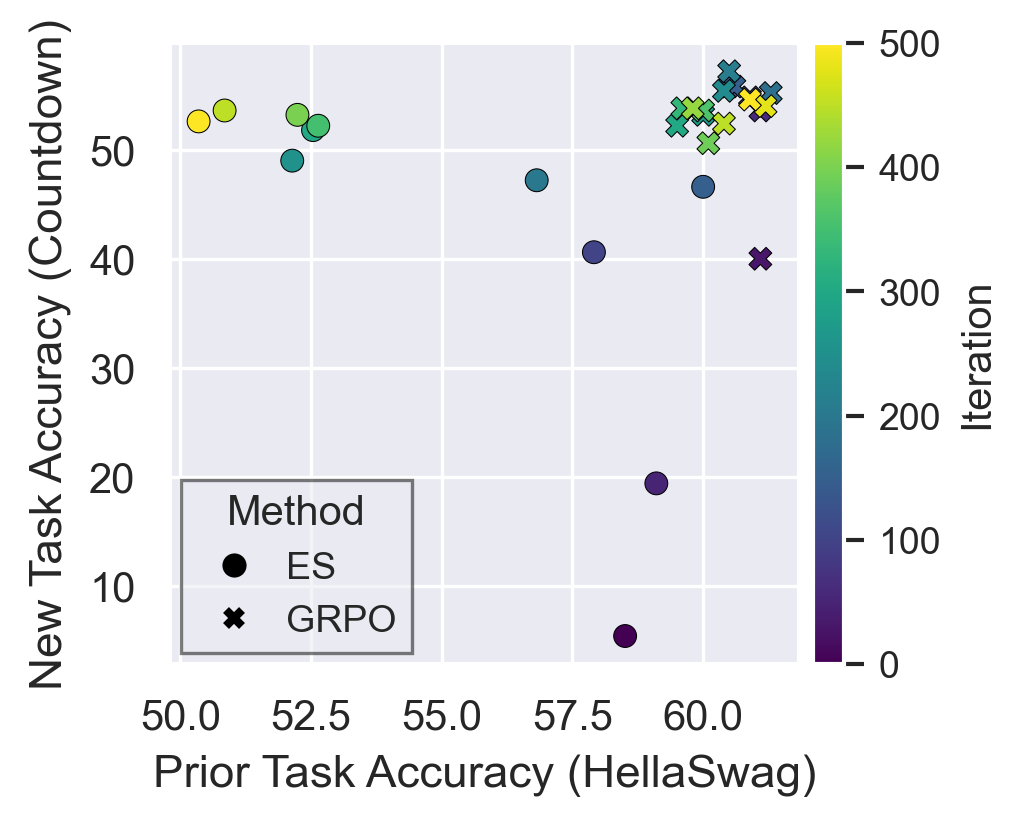}
\caption{Pareto front between new task (Countdown) and old task (HellaSwag) performance across fine-tuning with ES and GRPO.} 
\label{fig:new-vs-prev-task-performance-iter}
\end{figure}

\begin{figure}[t]
\centering
\includegraphics[width=0.8\columnwidth]{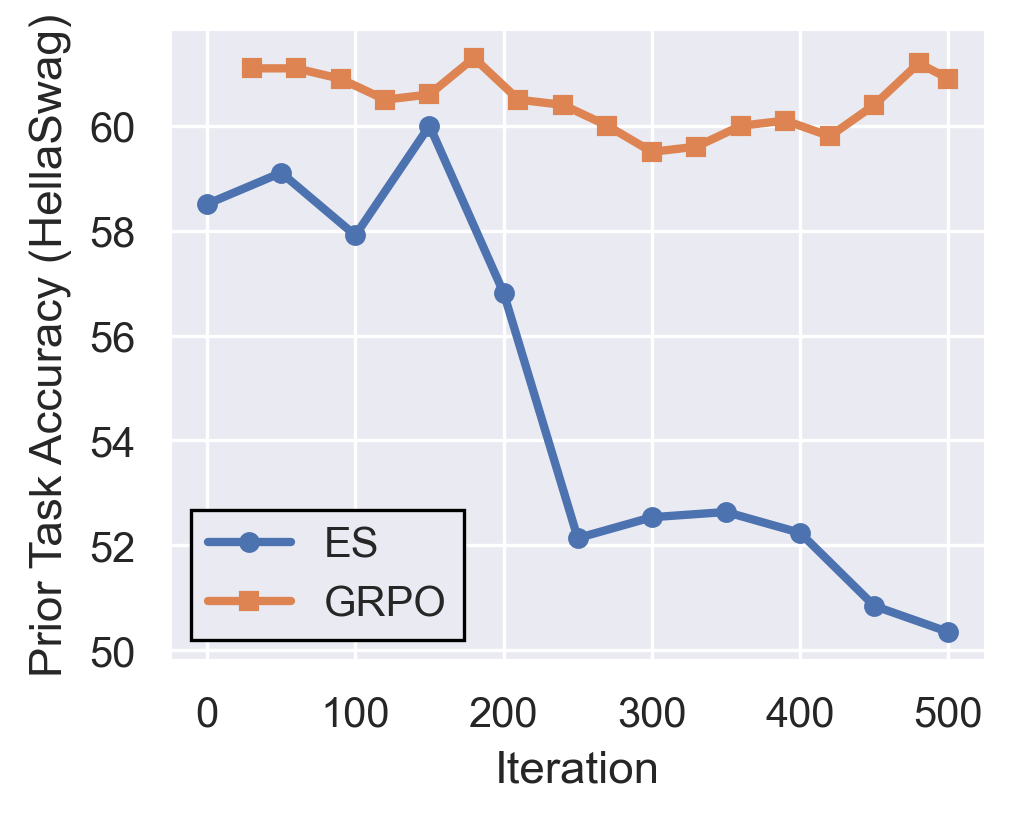}
\caption{Prior task accuracy (\%; HellaSwag) vs.\ training iteration for ES and GRPO fine-tuning. GRPO-trained models exhibit stable prior task accuracy across training iteration, while ES-trained models degrade with continued fine-tuning.}
\label{fig:hellaswag-vs-iteration}
\end{figure}

Figure~\ref{fig:new-vs-prev-task-performance-iter} illustrates the relationship between new-task performance (Countdown) and prior-task performance (HellaSwag) across fine-tuning iterations. When training with ES, prior-task performance systematically deteriorates as fine-tuning proceeds, even after new-task performance has effectively converged. This can observed in the convex Pareto front made by ES in Figure~\ref{fig:new-vs-prev-task-performance-iter}. The darker color dots, which depict early training iterations begin with a lower ``New Task Accuracy''. With enough training iterations, the increase in ``New Task Accuracy'' for ES is accompanied by a gradual but evident decline in ``Prior Task Accuracy''. Additionally, ES models reach near-maximum Countdown performance by approximately 200 iterations, after which additional training yields negligible gains on the new task. As shown in Figure \ref{fig:hellaswag-vs-iteration}, despite this convergence, previous task performance continues to decline with further iterations, resulting in an approximately 10\% drop relative to the best observed prior-task performance. \textbf{This pattern indicates that continued ES optimization disproportionately harms previously acquired capabilities, rather than trading off against improvements on the new task.}

In contrast, models fine-tuned with GRPO exhibit markedly different behavior. Across the full range of training iterations, GRPO maintains stable previous task performance while achieving strong new task accuracy. This can be seen by the cluster of crosses on the top-right corner of Figure~\ref{fig:new-vs-prev-task-performance-iter}. \textbf{This suggests that GRPO avoids the destructive interference observed with ES.} This property of GRPO has also been observed in prior work \cite{shenfeld2025rlsrazoronlinereinforcement}.

Therefore, we see that although ES-trained models can be competitive to GRPO, they do so at the cost of severe catastrophic forgetting. Notably, this forgetting occurs within a single fine-tuning run rather than across sequential tasks, highlighting a fundamental instability in ES-based continual adaptation. These results suggest that ES is poorly suited for scenarios requiring task generalization or reuse of previously learned capabilities, whereas GRPO provides a substantially more stable fine-tuning regime.

\begin{figure}[t]
\centering
\includegraphics[width=\linewidth]{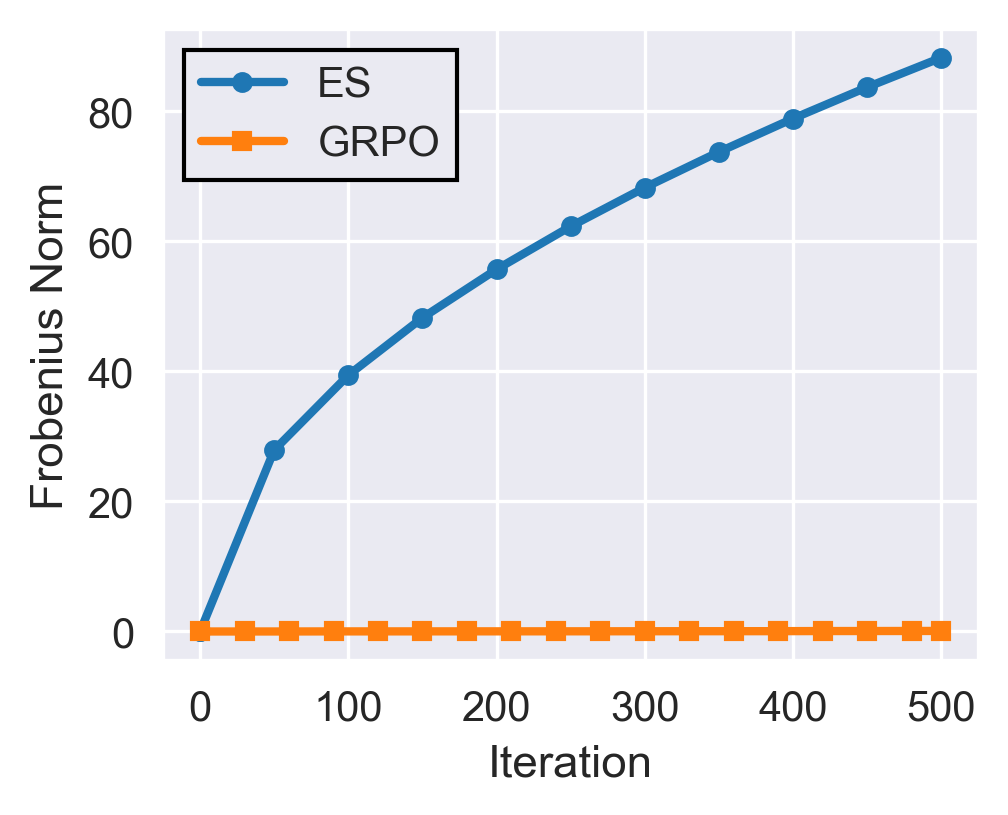}
\caption{
Relationship between Frobenius norm of a model update and number of training iterations on a new task.  ES-trained models drift several orders of magnitude more than GRPO-trained models.  
}

\label{fig:hellaswag-frobenius}
\end{figure}

\subsection{Dissecting ES Updates: Norm and Sparsity}

In this section, we seek to determine the characteristics of fine-tuning with ES that cause catastrophic forgetting. We do this by analyzing two features: the update norm and sparsity.

\paragraph{Norm.} Here we investigate norm growth of the updated matrix as a function of number of updates with ES and GRPO. We measure the Frobenius norm between model checkpoints within a training run. We do this for the Qwen2.5-1.5B-Instruct model trained on the Countdown dataset.

The results are shown in Figure~\ref{fig:hellaswag-frobenius}. The Frobenius norm increases monotonically with the number of training iterations for ES-trained models. A similar trend is also present for GRPO-trained models (Figure \ref{fig:hellaswag-frobenius-log}); however, the key distinction lies in scale. After just 500 training iteration, the Frobenius norm of the ES-trained model relative to the base model is \textit{three} orders of magnitude larger than the GRPO-trained model. When combined with what we learn from Figure \ref{fig:hellaswag-vs-iteration}, we see a clear association between the large increases in ES Frobenius norm and a decline in prior task accuracy. \textbf{Thus, ES updates have significantly higher $\ell_2$ norm difference, causing orders or magnitiude larger parameter-shifts compared to GRPO.}

\begin{figure}[t]
\centering
\setlength{\belowcaptionskip}{0pt}
\includegraphics[width=\linewidth]{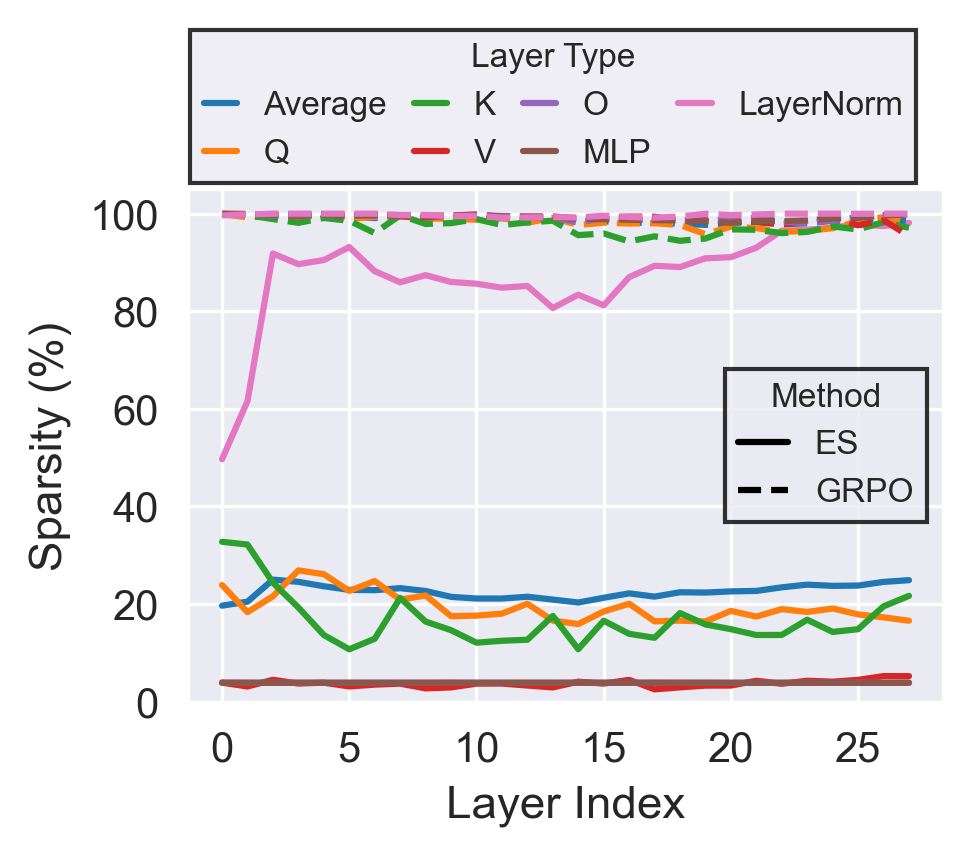}\\[-0.3em]
\vspace{-0.4em}
\caption{
Layerwise sparsity of parameter updates during fine-tuning (higher indicates more sparse updates).
ES exhibits broadly distributed, dense updates across components, whereas GRPO updates are overall highly sparse across LLM parameter groups, consistent with more targeted parameter changes.
}
\vspace{-0.6em}
\label{fig:update-sparsity}
\end{figure}

\paragraph{Sparsity.} Each update in ES is constructed from high-variance, global perturbations applied across all parameters, which may affect a large number of stored parameters uniformly. In contrast, it is known that GRPO applies much sparser and targeted updates via backpropagation, limiting the extent of unintended parameter drift \cite{mukherjee2025reinforcement}. To check the difference in the number of parameters affected by these algorithms, we evaluate the update sparsity in ES when compared to GRPO. 

We analyze Qwen2.5-1.5B-Instruct trained on the Countdown task with both GRPO and ES. We analyze the difference between a base model checkpoint and its corresponding fine-tuned checkpoint.  For each shared parameter tensor, we compute the update $\Delta W = W_{new} - W_{base}$. Following prior work \cite{mukherjee2025reinforcement}, we define sparsity as the percentage of elements whose absolute magnitude is below a fixed threshold ($\tau = 10^{-6}$). Therefore, higher sparsity values mean a larger number of parameters are below this threshold, which means that the updates are more sparse.  Parameters are grouped by architectural component, including attention projections $(Q, K, V)$, the attention output projection $(W_O)$, MLP layers and LayerNorms.  Updates are further aggregated by transformer layer index to obtain layerwise sparsity profiles.

The results are shown in Figure \ref{fig:update-sparsity}. \textbf{We see that ES updates are substantially less sparse across layers and parameter groups when compared to GRPO.} The sparsity levels for GRPO updates across all parameter types and layers are close to 95\%, which means updates are concentrated around a very small number of parameters. However, the updates using ES have very low sparsity levels, showing that a much larger number of parameters are perturbed when fine-tuning with ES. The most sparse updates in ES appear for LayerNorm; however it also contains the least (and neglible) number of parameters compared to other parts of the model. Other layer updates, irrespective of model depth, are highly dense in ES-trained models.  

Therefore, GRPO exhibits structured and comparatively sparse updates, aligning with the hypothesis that gradient-based optimization concentrates changes in task-relevant subspaces and mitigates interference with prior capabilities. When combined with KL regularization, these mechanisms provide a natural safeguard against large-scale parameter drift and, consequently, catastrophic forgetting.
In contrast, we see that updates using ES have orders of  magnitude larger norms and are much less sparse compared to GRPO. The lack of sparsity and large update norms in ES drifts the fine-tuned model further away from the base model, potentially leading to the catastrophic forgetting behavior observed in previous sections. 


\section{Conclusion}
We perform an empirical analysis of Evolutionary Strategies for fine-tuning LLMs based on recent work \cite{qiuEvolutionStrategiesScale2025} and show that it performs competetively with GRPO. Although, a critical roadblock still persists: \textbf{we observe that ES exhibits significant catastrophic forgetting and progressively deteriorates performance on prior skills of the model.} We show that this happens because ES updates have large norms and low sparsity levels (more dense), resulting in parameter drifts that are 1000x higher in magnitude than drifts observed with GRPO for the same number of update steps. These results imply that although recent progress in ES has bridged performance gap with state-of-the-art learning algorithms like GRPO, its intense model degradation still remains a challenge before its widespread adoption.

\section*{Limitations}
ES has an inherent randomness associated with the updates due to the nature of the algorithm. As a result, models trained with ES exhibit high variance in stochastic perturbations. Although we observed consistent qualitative trends across the settings we tested, where we worked with a population size of 30 as suggested in prior work \cite{qiuEvolutionStrategiesScale2025}, increased population size will decrease the variance and increase statistical stability of our performance numbers. 

Additionally, we evaluate catastrophic forgetting by tracking performance on one task during continued fine-tuning on Countdown, which measures retention of a broad prior capability. Doing so does not fully capture multi-facetted loss of performance that may be happening in the model; however, is enough to give strong evidence of the occurence of the phenomenon.

\bibliography{custom}

@misc{malladi2024finetuninglanguagemodelsjust,
      title={Fine-Tuning Language Models with Just Forward Passes}, 
      author={Sadhika Malladi and Tianyu Gao and Eshaan Nichani and Alex Damian and Jason D. Lee and Danqi Chen and Sanjeev Arora},
      year={2024},
      eprint={2305.17333},
      archivePrefix={arXiv},
      primaryClass={cs.LG},
      url={https://arxiv.org/abs/2305.17333}, 
}

@article{kirkpatrick2017overcoming,
  title={Overcoming catastrophic forgetting in neural networks},
  author={Kirkpatrick, James and Pascanu, Razvan and Rabinowitz, Neil and Veness, Joel and Desjardins, Guillaume and Rusu, Andrei A and Milan, Kieran and Quan, John and Ramalho, Tiago and Grabska-Barwinska, Agnieszka and others},
  journal={Proceedings of the national academy of sciences},
  volume={114},
  number={13},
  pages={3521--3526},
  year={2017},
  publisher={National Academy of Sciences}
}

@article{mukherjee2025reinforcement,
  title={Reinforcement Learning Finetunes Small Subnetworks in Large Language Models},
  author={Mukherjee, Sagnik and Yuan, Lifan and Hakkani-Tur, Dilek and Peng, Hao},
  journal={arXiv preprint arXiv:2505.11711},
  year={2025}
}

@article{gupta2024model,
  title={Model editing at scale leads to gradual and catastrophic forgetting},
  author={Gupta, Akshat and Rao, Anurag and Anumanchipalli, Gopala},
  journal={arXiv preprint arXiv:2401.07453},
  year={2024}
}

@misc{openai_memory_2024,
  title        = {Memory and New Controls for {ChatGPT}},
  author       = {{OpenAI}},
  year         = {2024},
  month        = {02},
  url          = {https://openai.com/index/memory-and-new-controls-for-chatgpt/},
  note         = {Accessed: 2026-01-24}
}

@misc{Sheng_2025, series={EuroSys ’25},
   title={HybridFlow: A Flexible and Efficient RLHF Framework},
   url={http://dx.doi.org/10.1145/3689031.3696075},
   DOI={10.1145/3689031.3696075},
   booktitle={Proceedings of the Twentieth European Conference on Computer Systems},
   publisher={ACM},
   author={Sheng, Guangming and Zhang, Chi and Ye, Zilingfeng and Wu, Xibin and Zhang, Wang and Zhang, Ru and Peng, Yanghua and Lin, Haibin and Wu, Chuan},
   year={2025},
   month=mar, pages={1279–1297},
   collection={EuroSys ’25} }

@misc{qiuEvolutionStrategiesScale2025,
  title = {Evolution {{Strategies}} at {{Scale}}: {{LLM Fine-Tuning Beyond Reinforcement Learning}}},
  shorttitle = {Evolution {{Strategies}} at {{Scale}}},
  author = {Qiu, Xin and Gan, Yulu and Hayes, Conor F. and Liang, Qiyao and Meyerson, Elliot and Hodjat, Babak and Miikkulainen, Risto},
  year = {2025},
  date = {2025-09-29},
  eprint = {2509.24372},
  eprinttype = {arXiv},
  eprintclass = {cs},
  doi = {10.48550/arXiv.2509.24372},
  url = {http://arxiv.org/abs/2509.24372},
  urldate = {2026-01-03},
  pubstate = {prepublished},
}

@misc{zellers2019hellaswagmachinereallyfinish,
      title={HellaSwag: Can a Machine Really Finish Your Sentence?}, 
      author={Rowan Zellers and Ari Holtzman and Yonatan Bisk and Ali Farhadi and Yejin Choi},
      year={2019},
      eprint={1905.07830},
      archivePrefix={arXiv},
      primaryClass={cs.CL},
      url={https://arxiv.org/abs/1905.07830}, 
}

@misc{shao2024deepseekmathpushinglimitsmathematical,
      title={DeepSeekMath: Pushing the Limits of Mathematical Reasoning in Open Language Models}, 
      author={Zhihong Shao and Peiyi Wang and Qihao Zhu and Runxin Xu and Junxiao Song and Xiao Bi and Haowei Zhang and Mingchuan Zhang and Y. K. Li and Y. Wu and Daya Guo},
      year={2024},
      eprint={2402.03300},
      archivePrefix={arXiv},
      primaryClass={cs.CL},
      url={https://arxiv.org/abs/2402.03300}, 
}

@article{hansen2001completelyderandomizedselfadaptationinevolutionstrategies,
    author = {Hansen, Nikolaus and Ostermeier, Andreas},
    title = {Completely Derandomized Self-Adaptation in Evolution Strategies},
    journal = {Evolutionary Computation},
    volume = {9},
    number = {2},
    pages = {159-195},
    year = {2001},
    month = {06},
    issn = {1063-6560},
    doi = {10.1162/106365601750190398},
    url = {https://doi.org/10.1162/106365601750190398},
    eprint = {https://direct.mit.edu/evco/article-pdf/9/2/159/1493523/106365601750190398.pdf},
}

@misc{korotyshova2025essaevolutionarystrategiesscalable,
      title={ESSA: Evolutionary Strategies for Scalable Alignment}, 
      author={Daria Korotyshova and Boris Shaposhnikov and Alexey Malakhov and Alexey Khokhulin and Nikita Surnachev and Kirill Ovcharenko and George Bredis and Alexey Gorbatovski and Viacheslav Sinii and Daniil Gavrilov},
      year={2025},
      eprint={2507.04453},
      archivePrefix={arXiv},
      primaryClass={cs.LG},
      url={https://arxiv.org/abs/2507.04453}, 
}

@misc{salimansEvolutionStrategiesScalable2017,
  title = {Evolution {{Strategies}} as a {{Scalable Alternative}} to {{Reinforcement Learning}}},
  author = {Salimans, Tim and Ho, Jonathan and Chen, Xi and Sidor, Szymon and Sutskever, Ilya},
  year = {2017},
  date = {2017},
  eprint = {1703.03864},
  eprinttype = {arXiv},
  eprintclass = {stat},
  doi = {10.48550/arXiv.1703.03864},
  url = {http://arxiv.org/abs/1703.03864},
  urldate = {2026-01-03},
  pubstate = {prepublished}
}

@misc{sarkar2025evolutionstrategieshyperscale,
      title={Evolution Strategies at the Hyperscale}, 
      author={Bidipta Sarkar and Mattie Fellows and Juan Agustin Duque and Alistair Letcher and Antonio León Villares and Anya Sims and Dylan Cope and Jarek Liesen and Lukas Seier and Theo Wolf and Uljad Berdica and Alexander David Goldie and Aaron Courville and Karin Sevegnani and Shimon Whiteson and Jakob Nicolaus Foerster},
      year={2025},
      eprint={2511.16652},
      archivePrefix={arXiv},
      primaryClass={cs.LG},
      url={https://arxiv.org/abs/2511.16652}, 
}

@misc{ouyangTrainingLanguageModels2022,
  title = {Training Language Models to Follow Instructions with Human Feedback},
  author = {Ouyang, Long and Wu, Jeff and Jiang, Xu and Almeida, Diogo and Wainwright, Carroll L. and Mishkin, Pamela and Zhang, Chong and Agarwal, Sandhini and Slama, Katarina and Ray, Alex and Schulman, John and Hilton, Jacob and Kelton, Fraser and Miller, Luke and Simens, Maddie and Askell, Amanda and Welinder, Peter and Christiano, Paul and Leike, Jan and Lowe, Ryan},
  date = {2022-03-04},
  year = {2022},
  eprint = {2203.02155},
  eprinttype = {arXiv},
  eprintclass = {cs},
  doi = {10.48550/arXiv.2203.02155},
  url = {http://arxiv.org/abs/2203.02155},
  urldate = {2026-01-03},
  pubstate = {prepublished}
}

@misc{cobbe2021trainingverifierssolvemath,
      title={Training Verifiers to Solve Math Word Problems}, 
      author={Karl Cobbe and Vineet Kosaraju and Mohammad Bavarian and Mark Chen and Heewoo Jun and Lukasz Kaiser and Matthias Plappert and Jerry Tworek and Jacob Hilton and Reiichiro Nakano and Christopher Hesse and John Schulman},
      year={2021},
      eprint={2110.14168},
      archivePrefix={arXiv},
      primaryClass={cs.LG},
      url={https://arxiv.org/abs/2110.14168}, 
}

@misc{hendrycks2021measuringmathematicalproblemsolving,
      title={Measuring Mathematical Problem Solving With the MATH Dataset}, 
      author={Dan Hendrycks and Collin Burns and Saurav Kadavath and Akul Arora and Steven Basart and Eric Tang and Dawn Song and Jacob Steinhardt},
      year={2021},
      eprint={2103.03874},
      archivePrefix={arXiv},
      primaryClass={cs.LG},
      url={https://arxiv.org/abs/2103.03874}, 
}

@misc{he2024olympiadbenchchallengingbenchmarkpromoting,
      title={OlympiadBench: A Challenging Benchmark for Promoting AGI with Olympiad-Level Bilingual Multimodal Scientific Problems}, 
      author={Chaoqun He and Renjie Luo and Yuzhuo Bai and Shengding Hu and Zhen Leng Thai and Junhao Shen and Jinyi Hu and Xu Han and Yujie Huang and Yuxiang Zhang and Jie Liu and Lei Qi and Zhiyuan Liu and Maosong Sun},
      year={2024},
      eprint={2402.14008},
      archivePrefix={arXiv},
      primaryClass={cs.CL},
      url={https://arxiv.org/abs/2402.14008}, 
}

@misc{deepseek-aiDeepSeekR1IncentivizingReasoning2025,
  title = {{{DeepSeek-R1}}: {{Incentivizing Reasoning Capability}} in {{LLMs}} via {{Reinforcement Learning}}},
  shorttitle = {{{DeepSeek-R1}}},
  year={2025},
  author = {DeepSeek-AI and Guo, Daya and Yang, Dejian and Zhang, Haowei and Song, Junxiao and Zhang, Ruoyu and Xu, Runxin and Zhu, Qihao and Ma, Shirong and Wang, Peiyi and Bi, Xiao and Zhang, Xiaokang and Yu, Xingkai and Wu, Yu and Wu, Z. F. and Gou, Zhibin and Shao, Zhihong and Li, Zhuoshu and Gao, Ziyi and Liu, Aixin and Xue, Bing and Wang, Bingxuan and Wu, Bochao and Feng, Bei and Lu, Chengda and Zhao, Chenggang and Deng, Chengqi and Zhang, Chenyu and Ruan, Chong and Dai, Damai and Chen, Deli and Ji, Dongjie and Li, Erhang and Lin, Fangyun and Dai, Fucong and Luo, Fuli and Hao, Guangbo and Chen, Guanting and Li, Guowei and Zhang, H. and Bao, Han and Xu, Hanwei and Wang, Haocheng and Ding, Honghui and Xin, Huajian and Gao, Huazuo and Qu, Hui and Li, Hui and Guo, Jianzhong and Li, Jiashi and Wang, Jiawei and Chen, Jingchang and Yuan, Jingyang and Qiu, Junjie and Li, Junlong and Cai, J. L. and Ni, Jiaqi and Liang, Jian and Chen, Jin and Dong, Kai and Hu, Kai and Gao, Kaige and Guan, Kang and Huang, Kexin and Yu, Kuai and Wang, Lean and Zhang, Lecong and Zhao, Liang and Wang, Litong and Zhang, Liyue and Xu, Lei and Xia, Leyi and Zhang, Mingchuan and Zhang, Minghua and Tang, Minghui and Li, Meng and Wang, Miaojun and Li, Mingming and Tian, Ning and Huang, Panpan and Zhang, Peng and Wang, Qiancheng and Chen, Qinyu and Du, Qiushi and Ge, Ruiqi and Zhang, Ruisong and Pan, Ruizhe and Wang, Runji and Chen, R. J. and Jin, R. L. and Chen, Ruyi and Lu, Shanghao and Zhou, Shangyan and Chen, Shanhuang and Ye, Shengfeng and Wang, Shiyu and Yu, Shuiping and Zhou, Shunfeng and Pan, Shuting and Li, S. S. and Zhou, Shuang and Wu, Shaoqing and Ye, Shengfeng and Yun, Tao and Pei, Tian and Sun, Tianyu and Wang, T. and Zeng, Wangding and Zhao, Wanjia and Liu, Wen and Liang, Wenfeng and Gao, Wenjun and Yu, Wenqin and Zhang, Wentao and Xiao, W. L. and An, Wei and Liu, Xiaodong and Wang, Xiaohan and Chen, Xiaokang and Nie, Xiaotao and Cheng, Xin and Liu, Xin and Xie, Xin and Liu, Xingchao and Yang, Xinyu and Li, Xinyuan and Su, Xuecheng and Lin, Xuheng and Li, X. Q. and Jin, Xiangyue and Shen, Xiaojin and Chen, Xiaosha and Sun, Xiaowen and Wang, Xiaoxiang and Song, Xinnan and Zhou, Xinyi and Wang, Xianzu and Shan, Xinxia and Li, Y. K. and Wang, Y. Q. and Wei, Y. X. and Zhang, Yang and Xu, Yanhong and Li, Yao and Zhao, Yao and Sun, Yaofeng and Wang, Yaohui and Yu, Yi and Zhang, Yichao and Shi, Yifan and Xiong, Yiliang and He, Ying and Piao, Yishi and Wang, Yisong and Tan, Yixuan and Ma, Yiyang and Liu, Yiyuan and Guo, Yongqiang and Ou, Yuan and Wang, Yuduan and Gong, Yue and Zou, Yuheng and He, Yujia and Xiong, Yunfan and Luo, Yuxiang and You, Yuxiang and Liu, Yuxuan and Zhou, Yuyang and Zhu, Y. X. and Xu, Yanhong and Huang, Yanping and Li, Yaohui and Zheng, Yi and Zhu, Yuchen and Ma, Yunxian and Tang, Ying and Zha, Yukun and Yan, Yuting and Ren, Z. Z. and Ren, Zehui and Sha, Zhangli and Fu, Zhe and Xu, Zhean and Xie, Zhenda and Zhang, Zhengyan and Hao, Zhewen and Ma, Zhicheng and Yan, Zhigang and Wu, Zhiyu and Gu, Zihui and Zhu, Zijia and Liu, Zijun and Li, Zilin and Xie, Ziwei and Song, Ziyang and Pan, Zizheng and Huang, Zhen and Xu, Zhipeng and Zhang, Zhongyu and Zhang, Zhen},
  date = {2025-01-22},
  eprint = {2501.12948},
  eprinttype = {arXiv},
  eprintclass = {cs},
  doi = {10.48550/arXiv.2501.12948},
  url = {http://arxiv.org/abs/2501.12948},
  urldate = {2026-01-03},
  pubstate = {prepublished}
}

@InProceedings{10.1007/978-3-642-83814-9_6,
author="Rechenberg, Ingo",
editor="Bergmann, H. W.",
title="Evolution Strategy: Nature's Way of Optimization",
booktitle="Optimization: Methods and Applications, Possibilities and Limitations",
year="1989",
publisher="Springer Berlin Heidelberg",
address="Berlin, Heidelberg",
pages="106--126",
isbn="978-3-642-83814-9"
}

@book{schwefel1977numerische,
  author = {Schwefel, Hans-Paul},
  title = {Numerische Optimierung von Computer-Modellen mittels der Evolutionsstrategie: mit einer vergleichenden Einf{\"u}hrung in die Hill-Climbing- und Zufallsstrategie},
  year = {1977},
  publisher = {Birkh{\"a}user Verlag},
  address = {Basel, Stuttgart},
  isbn = {978-3-7643-0926-8}
}

@Article{Beyer1995b,
  author = {Beyer, H.-G.},
  title = {Toward a theory of evolution strategies: the ($\mu$, $\lambda$)-theory},
  journal = {Evolutionary Computation},
  volume = {2},
  number = {4},
  pages = {381--407},
  year = {1995},
  doi = {10.1162/evco.1995.2.4.381},
  publisher = {MIT Press}
}

@misc{wierstra2011naturalevolutionstrategies,
      title={Natural Evolution Strategies}, 
      author={Daan Wierstra and Tom Schaul and Tobias Glasmachers and Yi Sun and Jürgen Schmidhuber},
      year={2011},
      eprint={1106.4487},
      archivePrefix={arXiv},
      primaryClass={stat.ML},
      url={https://arxiv.org/abs/1106.4487}, 
}

@misc{sun2012efficientnaturalevolutionstrategies,
      title={Efficient Natural Evolution Strategies}, 
      author={Yi Sun and Daan Wierstra and Tom Schaul and Juergen Schmidhuber},
      year={2012},
      eprint={1209.5853},
      archivePrefix={arXiv},
      primaryClass={cs.AI},
      url={https://arxiv.org/abs/1209.5853}, 
}

@misc{zhang2017relationshipopenaievolutionstrategy,
      title={On the Relationship Between the OpenAI Evolution Strategy and Stochastic Gradient Descent}, 
      author={Xingwen Zhang and Jeff Clune and Kenneth O. Stanley},
      year={2017},
      eprint={1712.06564},
      archivePrefix={arXiv},
      primaryClass={cs.NE},
      url={https://arxiv.org/abs/1712.06564}, 
}

@misc{risi2019deepneuroevolutionrecurrentdiscrete,
      title={Deep Neuroevolution of Recurrent and Discrete World Models}, 
      author={Sebastian Risi and Kenneth O. Stanley},
      year={2019},
      eprint={1906.08857},
      archivePrefix={arXiv},
      primaryClass={cs.NE},
      url={https://arxiv.org/abs/1906.08857}, 
}

@misc{such2018deepneuroevolutiongeneticalgorithms,
      title={Deep Neuroevolution: Genetic Algorithms Are a Competitive Alternative for Training Deep Neural Networks for Reinforcement Learning}, 
      author={Felipe Petroski Such and Vashisht Madhavan and Edoardo Conti and Joel Lehman and Kenneth O. Stanley and Jeff Clune},
      year={2018},
      eprint={1712.06567},
      archivePrefix={arXiv},
      primaryClass={cs.NE},
      url={https://arxiv.org/abs/1712.06567}, 
}

@misc{jin2024derivativefreeoptimizationlowrankadaptation,
      title={Derivative-Free Optimization for Low-Rank Adaptation in Large Language Models}, 
      author={Feihu Jin and Yin Liu and Ying Tan},
      year={2024},
      eprint={2403.01754},
      archivePrefix={arXiv},
      primaryClass={cs.CL},
      url={https://arxiv.org/abs/2403.01754}, 
}

@article{vaswani2017attention,
  title   = {Attention Is All You Need},
  author  = {Vaswani, Ashish and Shazeer, Noam and Parmar, Niki and Uszkoreit, Jakob and Jones, Llion and Gomez, Aidan N. and Kaiser, Lukasz and Polosukhin, Illia},
  journal = {arXiv preprint arXiv:1706.03762},
  year    = {2017},
  doi     = {10.48550/arXiv.1706.03762},
  url     = {https://arxiv.org/abs/1706.03762}
}

@article{brown2020language,
  title   = {Language Models are Few-Shot Learners},
  author  = {Brown, Tom B. and Mann, Benjamin and Ryder, Nick and Subbiah, Melanie and Kaplan, Jared and Dhariwal, Prafulla and Neelakantan, Arvind and Shyam, Pranav and Sastry, Girish and Askell, Amanda and Agarwal, Sandhini and Herbert-Voss, Ariel and Krueger, Gretchen and Henighan, Tom and Child, Rewon and Ramesh, Aditya and Ziegler, Daniel M. and Wu, Jeffrey and Winter, Clemens and Hesse, Christopher and Chen, Mark and Sigler, Eric and Litwin, Mateusz and Gray, Scott and Chess, Benjamin and Clark, Jack and Berner, Christopher and McCandlish, Sam and Radford, Alec and Sutskever, Ilya and Amodei, Dario},
  journal = {arXiv preprint arXiv:2005.14165},
  year    = {2020},
  doi     = {10.48550/arXiv.2005.14165},
  url     = {https://arxiv.org/abs/2005.14165}
}

@article{deepseekai2024deepseekllm,
  title   = {DeepSeek LLM: Scaling Open-Source Language Models with Longtermism},
  author  = {{DeepSeek-AI} and Bi, Xiao and Chen, Deli and Chen, Guanting and Chen, Shanhuang and Dai, Damai and Deng, Chengqi and Ding, Honghui and Dong, Kai and Du, Qiushi and Fu, Zhe and Gao, Huazuo and Gao, Kaige and Gao, Wenjun and Ge, Ruiqi and Guan, Kang and Guo, Daya and Guo, Jianzhong and Hao, Guangbo and Hao, Zhewen and He, Ying and Hu, Wenjie and Huang, Panpan and Li, Erhang and Li, Guowei and Li, Jiashi and Li, Yao and Liang, Wenfeng and Lin, Fangyun and Liu, A. X. and Liu, Bo and Liu, Wen and Liu, Xiaodong and Liu, Xin and Liu, Yiyuan and Lu, Haoyu and Lu, Shanghao and Luo, Fuli and Ma, Shirong and Nie, Xiaotao and Pei, Tian and Piao, Yishi and Qiu, Junjie and Qu, Hui and Ren, Tongzheng and Ren, Zehui and Ruan, Chong and Sha, Zhangli and Shao, Zhihong and Song, Junxiao and Su, Xuecheng and Sun, Jingxiang and Sun, Yaofeng and Tang, Minghui and Wang, Bingxuan and Wang, Peiyi and Wang, Shiyu and Wang, Yaohui and Wu, Tong and Xie, Xin and Xiong, Yiliang and Xu, Hanwei and Yang, Dejian and You, Yuxiang and Yu, Shuiping and Yu, Xingkai and Zhang, B. and Zhang, Haowei and Zhang, Lecong and Zhang, Minghua and Zhao, Chenggang and Zhao, Yao and Zhou, Shunfeng and Zhu, Qihao and Zou, Yuheng and others},
  journal = {arXiv preprint arXiv:2401.02954},
  year    = {2024},
  doi     = {10.48550/arXiv.2401.02954},
  url     = {https://arxiv.org/abs/2401.02954}
}

@misc{rafailov2024directpreferenceoptimizationlanguage,
      title={Direct Preference Optimization: Your Language Model is Secretly a Reward Model}, 
      author={Rafael Rafailov and Archit Sharma and Eric Mitchell and Stefano Ermon and Christopher D. Manning and Chelsea Finn},
      year={2024},
      eprint={2305.18290},
      archivePrefix={arXiv},
      primaryClass={cs.LG},
      url={https://arxiv.org/abs/2305.18290}, 
}

@misc{panJiayiPanTinyZero2026,
  title = {Jiayi-{{Pan}}/{{TinyZero}}},
  year = {2026},
  author = {Pan, Jiayi},
  date = {2026-01-03T00:19:41Z},
  origdate = {2025-01-21T16:49:12Z},
  url = {https://github.com/Jiayi-Pan/TinyZero},
  urldate = {2026-01-03}
}

@misc{wei2022finetunedlanguagemodelszeroshot,
      title={Finetuned Language Models Are Zero-Shot Learners}, 
      author={Jason Wei and Maarten Bosma and Vincent Y. Zhao and Kelvin Guu and Adams Wei Yu and Brian Lester and Nan Du and Andrew M. Dai and Quoc V. Le},
      year={2022},
      eprint={2109.01652},
      archivePrefix={arXiv},
      primaryClass={cs.CL},
      url={https://arxiv.org/abs/2109.01652}, 
}

@misc{grattafiori2024llama3herdmodels,
      title={The Llama 3 Herd of Models}, 
      author={Aaron Grattafiori and Abhimanyu Dubey and Abhinav Jauhri and Abhinav Pandey and Abhishek Kadian and Ahmad Al-Dahle and Aiesha Letman and Akhil Mathur and Alan Schelten and Alex Vaughan and Amy Yang and Angela Fan and Anirudh Goyal and Anthony Hartshorn and Aobo Yang and Archi Mitra and Archie Sravankumar and Artem Korenev and Arthur Hinsvark and Arun Rao and Aston Zhang and Aurelien Rodriguez and Austen Gregerson and Ava Spataru and Baptiste Roziere and Bethany Biron and Binh Tang and Bobbie Chern and Charlotte Caucheteux and Chaya Nayak and Chloe Bi and Chris Marra and Chris McConnell and Christian Keller and Christophe Touret and Chunyang Wu and Corinne Wong and Cristian Canton Ferrer and Cyrus Nikolaidis and Damien Allonsius and Daniel Song and Danielle Pintz and Danny Livshits and Danny Wyatt and David Esiobu and Dhruv Choudhary and Dhruv Mahajan and Diego Garcia-Olano and Diego Perino and Dieuwke Hupkes and Egor Lakomkin and Ehab AlBadawy and Elina Lobanova and Emily Dinan and Eric Michael Smith and Filip Radenovic and Francisco Guzmán and Frank Zhang and Gabriel Synnaeve and Gabrielle Lee and Georgia Lewis Anderson and Govind Thattai and Graeme Nail and Gregoire Mialon and Guan Pang and Guillem Cucurell and Hailey Nguyen and Hannah Korevaar and Hu Xu and Hugo Touvron and Iliyan Zarov and Imanol Arrieta Ibarra and Isabel Kloumann and Ishan Misra and Ivan Evtimov and Jack Zhang and Jade Copet and Jaewon Lee and Jan Geffert and Jana Vranes and Jason Park and Jay Mahadeokar and Jeet Shah and Jelmer van der Linde and Jennifer Billock and Jenny Hong and Jenya Lee and Jeremy Fu and Jianfeng Chi and Jianyu Huang and Jiawen Liu and Jie Wang and Jiecao Yu and Joanna Bitton and Joe Spisak and Jongsoo Park and Joseph Rocca and Joshua Johnstun and Joshua Saxe and Junteng Jia and Kalyan Vasuden Alwala and Karthik Prasad and Kartikeya Upasani and Kate Plawiak and Ke Li and Kenneth Heafield and Kevin Stone and Khalid El-Arini and Krithika Iyer and Kshitiz Malik and Kuenley Chiu and Kunal Bhalla and Kushal Lakhotia and Lauren Rantala-Yeary and Laurens van der Maaten and Lawrence Chen and Liang Tan and Liz Jenkins and Louis Martin and Lovish Madaan and Lubo Malo and Lukas Blecher and Lukas Landzaat and Luke de Oliveira and Madeline Muzzi and Mahesh Pasupuleti and Mannat Singh and Manohar Paluri and Marcin Kardas and Maria Tsimpoukelli and Mathew Oldham and Mathieu Rita and Maya Pavlova and Melanie Kambadur and Mike Lewis and Min Si and Mitesh Kumar Singh and Mona Hassan and Naman Goyal and Narjes Torabi and Nikolay Bashlykov and Nikolay Bogoychev and Niladri Chatterji and Ning Zhang and Olivier Duchenne and Onur Çelebi and Patrick Alrassy and Pengchuan Zhang and Pengwei Li and Petar Vasic and Peter Weng and Prajjwal Bhargava and Pratik Dubal and Praveen Krishnan and Punit Singh Koura and Puxin Xu and Qing He and Qingxiao Dong and Ragavan Srinivasan and Raj Ganapathy and Ramon Calderer and Ricardo Silveira Cabral and Robert Stojnic and Roberta Raileanu and Rohan Maheswari and Rohit Girdhar and Rohit Patel and Romain Sauvestre and Ronnie Polidoro and Roshan Sumbaly and Ross Taylor and Ruan Silva and Rui Hou and Rui Wang and Saghar Hosseini and Sahana Chennabasappa and Sanjay Singh and Sean Bell and Seohyun Sonia Kim and Sergey Edunov and Shaoliang Nie and Sharan Narang and Sharath Raparthy and Sheng Shen and Shengye Wan and Shruti Bhosale and Shun Zhang and Simon Vandenhende and Soumya Batra and Spencer Whitman and Sten Sootla and Stephane Collot and Suchin Gururangan and Sydney Borodinsky and Tamar Herman and Tara Fowler and Tarek Sheasha and Thomas Georgiou and Thomas Scialom and Tobias Speckbacher and Todor Mihaylov and Tong Xiao and Ujjwal Karn and Vedanuj Goswami and Vibhor Gupta and Vignesh Ramanathan and Viktor Kerkez and Vincent Gonguet and Virginie Do and Vish Vogeti and Vítor Albiero and Vladan Petrovic and Weiwei Chu and Wenhan Xiong and Wenyin Fu and Whitney Meers and Xavier Martinet and Xiaodong Wang and Xiaofang Wang and Xiaoqing Ellen Tan and Xide Xia and Xinfeng Xie and Xuchao Jia and Xuewei Wang and Yaelle Goldschlag and Yashesh Gaur and Yasmine Babaei and Yi Wen and Yiwen Song and Yuchen Zhang and Yue Li and Yuning Mao and Zacharie Delpierre Coudert and Zheng Yan and Zhengxing Chen and Zoe Papakipos and Aaditya Singh and Aayushi Srivastava and Abha Jain and Adam Kelsey and Adam Shajnfeld and Adithya Gangidi and Adolfo Victoria and Ahuva Goldstand and Ajay Menon and Ajay Sharma and Alex Boesenberg and Alexei Baevski and Allie Feinstein and Amanda Kallet and Amit Sangani and Amos Teo and Anam Yunus and Andrei Lupu and Andres Alvarado and Andrew Caples and Andrew Gu and Andrew Ho and Andrew Poulton and Andrew Ryan and Ankit Ramchandani and Annie Dong and Annie Franco and Anuj Goyal and Aparajita Saraf and Arkabandhu Chowdhury and Ashley Gabriel and Ashwin Bharambe and Assaf Eisenman and Azadeh Yazdan and Beau James and Ben Maurer and Benjamin Leonhardi and Bernie Huang and Beth Loyd and Beto De Paola and Bhargavi Paranjape and Bing Liu and Bo Wu and Boyu Ni and Braden Hancock and Bram Wasti and Brandon Spence and Brani Stojkovic and Brian Gamido and Britt Montalvo and Carl Parker and Carly Burton and Catalina Mejia and Ce Liu and Changhan Wang and Changkyu Kim and Chao Zhou and Chester Hu and Ching-Hsiang Chu and Chris Cai and Chris Tindal and Christoph Feichtenhofer and Cynthia Gao and Damon Civin and Dana Beaty and Daniel Kreymer and Daniel Li and David Adkins and David Xu and Davide Testuggine and Delia David and Devi Parikh and Diana Liskovich and Didem Foss and Dingkang Wang and Duc Le and Dustin Holland and Edward Dowling and Eissa Jamil and Elaine Montgomery and Eleonora Presani and Emily Hahn and Emily Wood and Eric-Tuan Le and Erik Brinkman and Esteban Arcaute and Evan Dunbar and Evan Smothers and Fei Sun and Felix Kreuk and Feng Tian and Filippos Kokkinos and Firat Ozgenel and Francesco Caggioni and Frank Kanayet and Frank Seide and Gabriela Medina Florez and Gabriella Schwarz and Gada Badeer and Georgia Swee and Gil Halpern and Grant Herman and Grigory Sizov and Guangyi and Zhang and Guna Lakshminarayanan and Hakan Inan and Hamid Shojanazeri and Han Zou and Hannah Wang and Hanwen Zha and Haroun Habeeb and Harrison Rudolph and Helen Suk and Henry Aspegren and Hunter Goldman and Hongyuan Zhan and Ibrahim Damlaj and Igor Molybog and Igor Tufanov and Ilias Leontiadis and Irina-Elena Veliche and Itai Gat and Jake Weissman and James Geboski and James Kohli and Janice Lam and Japhet Asher and Jean-Baptiste Gaya and Jeff Marcus and Jeff Tang and Jennifer Chan and Jenny Zhen and Jeremy Reizenstein and Jeremy Teboul and Jessica Zhong and Jian Jin and Jingyi Yang and Joe Cummings and Jon Carvill and Jon Shepard and Jonathan McPhie and Jonathan Torres and Josh Ginsburg and Junjie Wang and Kai Wu and Kam Hou U and Karan Saxena and Kartikay Khandelwal and Katayoun Zand and Kathy Matosich and Kaushik Veeraraghavan and Kelly Michelena and Keqian Li and Kiran Jagadeesh and Kun Huang and Kunal Chawla and Kyle Huang and Lailin Chen and Lakshya Garg and Lavender A and Leandro Silva and Lee Bell and Lei Zhang and Liangpeng Guo and Licheng Yu and Liron Moshkovich and Luca Wehrstedt and Madian Khabsa and Manav Avalani and Manish Bhatt and Martynas Mankus and Matan Hasson and Matthew Lennie and Matthias Reso and Maxim Groshev and Maxim Naumov and Maya Lathi and Meghan Keneally and Miao Liu and Michael L. Seltzer and Michal Valko and Michelle Restrepo and Mihir Patel and Mik Vyatskov and Mikayel Samvelyan and Mike Clark and Mike Macey and Mike Wang and Miquel Jubert Hermoso and Mo Metanat and Mohammad Rastegari and Munish Bansal and Nandhini Santhanam and Natascha Parks and Natasha White and Navyata Bawa and Nayan Singhal and Nick Egebo and Nicolas Usunier and Nikhil Mehta and Nikolay Pavlovich Laptev and Ning Dong and Norman Cheng and Oleg Chernoguz and Olivia Hart and Omkar Salpekar and Ozlem Kalinli and Parkin Kent and Parth Parekh and Paul Saab and Pavan Balaji and Pedro Rittner and Philip Bontrager and Pierre Roux and Piotr Dollar and Polina Zvyagina and Prashant Ratanchandani and Pritish Yuvraj and Qian Liang and Rachad Alao and Rachel Rodriguez and Rafi Ayub and Raghotham Murthy and Raghu Nayani and Rahul Mitra and Rangaprabhu Parthasarathy and Raymond Li and Rebekkah Hogan and Robin Battey and Rocky Wang and Russ Howes and Ruty Rinott and Sachin Mehta and Sachin Siby and Sai Jayesh Bondu and Samyak Datta and Sara Chugh and Sara Hunt and Sargun Dhillon and Sasha Sidorov and Satadru Pan and Saurabh Mahajan and Saurabh Verma and Seiji Yamamoto and Sharadh Ramaswamy and Shaun Lindsay and Shaun Lindsay and Sheng Feng and Shenghao Lin and Shengxin Cindy Zha and Shishir Patil and Shiva Shankar and Shuqiang Zhang and Shuqiang Zhang and Sinong Wang and Sneha Agarwal and Soji Sajuyigbe and Soumith Chintala and Stephanie Max and Stephen Chen and Steve Kehoe and Steve Satterfield and Sudarshan Govindaprasad and Sumit Gupta and Summer Deng and Sungmin Cho and Sunny Virk and Suraj Subramanian and Sy Choudhury and Sydney Goldman and Tal Remez and Tamar Glaser and Tamara Best and Thilo Koehler and Thomas Robinson and Tianhe Li and Tianjun Zhang and Tim Matthews and Timothy Chou and Tzook Shaked and Varun Vontimitta and Victoria Ajayi and Victoria Montanez and Vijai Mohan and Vinay Satish Kumar and Vishal Mangla and Vlad Ionescu and Vlad Poenaru and Vlad Tiberiu Mihailescu and Vladimir Ivanov and Wei Li and Wenchen Wang and Wenwen Jiang and Wes Bouaziz and Will Constable and Xiaocheng Tang and Xiaojian Wu and Xiaolan Wang and Xilun Wu and Xinbo Gao and Yaniv Kleinman and Yanjun Chen and Ye Hu and Ye Jia and Ye Qi and Yenda Li and Yilin Zhang and Ying Zhang and Yossi Adi and Youngjin Nam and Yu and Wang and Yu Zhao and Yuchen Hao and Yundi Qian and Yunlu Li and Yuzi He and Zach Rait and Zachary DeVito and Zef Rosnbrick and Zhaoduo Wen and Zhenyu Yang and Zhiwei Zhao and Zhiyu Ma},
      year={2024},
      eprint={2407.21783},
      archivePrefix={arXiv},
      primaryClass={cs.AI},
      url={https://arxiv.org/abs/2407.21783}, 
}

@misc{qwenQwen25TechnicalReport2024,
  title = {Qwen2.5 {{Technical Report}}},
  year = {2024},
  author = {Qwen and Yang, An and Yang, Baosong and Zhang, Beichen and Hui, Binyuan and Zheng, Bo and Yu, Bowen and Li, Chengyuan and Liu, Dayiheng and Huang, Fei and Wei, Haoran and Lin, Huan and Yang, Jian and Tu, Jianhong and Zhang, Jianwei and Yang, Jianxin and Yang, Jiaxi and Zhou, Jingren and Lin, Junyang and Dang, Kai and Lu, Keming and Bao, Keqin and Yang, Kexin and Yu, Le and Li, Mei and Xue, Mingfeng and Zhang, Pei and Zhu, Qin and Men, Rui and Lin, Runji and Li, Tianhao and Tang, Tianyi and Xia, Tingyu and Ren, Xingzhang and Ren, Xuancheng and Fan, Yang and Su, Yang and Zhang, Yichang and Wan, Yu and Liu, Yuqiong and Cui, Zeyu and Zhang, Zhenru and Qiu, Zihan},
  date = {2024-12-19},
  url = {https://arxiv.org/abs/2412.15115v2},
  urldate = {2026-01-03},
  langid = {english},
  organization = {arXiv.org}
}

@misc{shenfeld2025rlsrazoronlinereinforcement,
      title={RL's Razor: Why Online Reinforcement Learning Forgets Less}, 
      author={Idan Shenfeld and Jyothish Pari and Pulkit Agrawal},
      year={2025},
      eprint={2509.04259},
      archivePrefix={arXiv},
      primaryClass={cs.LG},
      url={https://arxiv.org/abs/2509.04259}, 
}

\appendix
\section{Appendix} \label{sec:appendix}
\setcounter{figure}{0}
\renewcommand{\thefigure}{A\arabic{figure}}
\subsection{Algorithmic Overview and Analogy Between ES and GRPO} \label{appendix:A.1}

\subsubsection{ES Algorithm Overview}
\citet{qiuEvolutionStrategiesScale2025} implement a version of evolutionary strategies that features these techniques: weight adjustment in-place with noise generation from stored random seeds, ranked weight updates, and learning rate ingestion. 

Each update step can be understood through the following equations. 

Each population member at step $t$ has a unique seed. With noise per iteration $\epsilon_{n,l} \sim \mathcal{N}(0, I)$, model parameters for timestep  $t$ $\theta_t$, layer parameters for step $t$ $\theta_{t,l} $, reward function $R$, reward score for the $n$th member $R_n$, z-score for $n$th member $Z_n$, and noise coefficient $\sigma$, and learning rate $\alpha$. 

Reset random seed generator. Sample noise $\epsilon_{n,l} \sim \mathcal{N}(0, I)$. For all layers, perturb in-place:
\[
                \theta_{t-1,l} \leftarrow \theta_{t-1,l} + \sigma \cdot \epsilon_{n,l}.
\]

Reward for perturbed model is calculated:
\[
R_n = R(\theta_{t-1}).
\]

Reset random seed generator. Sample noise $\epsilon_{n,l} \sim \mathcal{N}(0, I)$. For all layers, restore in-place:
\[
                \theta_{t-1,l} \leftarrow \theta_{t-1,l} - \sigma \cdot \epsilon_{n,l}.
            \]

Z-score is calculated per population member: 
\[
        Z_n = \frac{R_n - R_{\text{mean}}}{R_{\text{std}}},
    \]

Reset random seed generator. Sample noise $\epsilon_{n,l} \sim \mathcal{N}(0, I)$. For all layers, update with noise weighted by z-score and learning rate in-place:
\[
                \theta_{t,l} \leftarrow \theta_{t-1,l} + \alpha \cdot \frac{1}{N} Z_n \epsilon_{n,l}.
            \]
            
    where $R_{\text{mean}}$ and $R_{\text{std}}$ are the mean and standard deviation of $R_1, R_2, \dots, R_N$.

\subsubsection{ES Algorithm Overview}
\citet{shao2024deepseekmathpushinglimitsmathematical} implement Group Relative Policy Optimization (GRPO), which eliminates the critic model by estimating advantages from group statistics. 

For each prompt $q$, sample a group of $G$ outputs $\{o_1, o_2, \dots, o_G\}$ from the current policy $\pi_{\theta_{old}}$.

Compute rewards $\{r_1, r_2, \dots, r_G\}$ for each output and calculate relative advantages via z-score normalization:
\[
    A_i = \frac{r_i - \text{mean}(\{r_1, \dots, r_G\})}{\text{std}(\{r_1, \dots, r_G\})}.
\]

The policy $\pi_\theta$ is updated by maximizing the GRPO objective:

\[
\rho_i(\theta) = \frac{\pi_\theta(o_i \mid q)}{\pi_{\theta_{\text{old}}}(o_i \mid q)}
\]

\[
\begin{aligned}
\mathcal{J}_{\text{GRPO}}(\theta)
&= \frac{1}{G} \sum_{i=1}^{G}
\min \Big(
\rho_i(\theta) A_i,\;\\
&
\text{clip}(\rho_i(\theta), 1-\epsilon, 1+\epsilon) A_i
\Big)\\
& - \beta \, \mathbb{D}_{\mathrm{KL}}(\pi_\theta \,\|\, \pi_{\text{ref}}).
\end{aligned}
\]

To penalize divergence from the reference policy $\pi_{ref}$ without additional sampling, the KL term is approximated:
\[
    \mathbb{D}_{KL}(\pi_\theta || \pi_{ref}) = \frac{\pi_{ref}(o_i|q)}{\pi_\theta(o_i|q)} - \log \frac{\pi_{ref}(o_i|q)}{\pi_\theta(o_i|q)} - 1.
\]

By replacing the value function with group-relative rewards, this implementation reduces computational overhead and memory usage compared to standard PPO.

\subsubsection{Analogy Between ES Population Size and GRPO Rollout Count}

Both GRPO and ES rely on creating different responses and then updating the model parameters via the fitness of those responses. The following section describes why the population size in ES and number of rollouts in GRPO play an analogous role in controlling parameter updates. 

Following the algorithm described by \citet{qiuEvolutionStrategiesScale2025}, an ES training update comprises of $N$ different seeds used to generate perturbations to the baseline model, resulting in $N$ different population members. Each population member is sampled with temperature at 0 to generate $N$ different responses, which are evaluated by a reward function to determine their fitness, which is converted into a z-score to weight the contribution of that respective perturbation to the baseline model. Further explanation can be found in \ref{appendix:A.1}. 

Similarly, a GRPO training update samples $N$ candidate outputs from the current policy, evaluates them to obtain relative reward signals, and updates the policy via a policy-gradient objective while constraining deviation from a fixed reference policy through KL regularization \citep{deepseek-aiDeepSeekR1IncentivizingReasoning2025}. Crucially, although ES simultaneously maintains multiple different versions of a model and GRPO maintains one, ES population size and GRPO number of rollouts both determine the number of samples used to estimate a stochastic update and to form a stochastic gradient or gradient-free estimator that drives the parameter update.

\subsection{Implementation Details} \label{appendix:A.4}
Our implementations for both GRPO and ES model training and analysis is attached to this submission.
\subsubsection{GRPO}
The GRPO setup in this study is implemented on the VERL library, which employs the HybridFlow engine proposed by \citet{Sheng_2025}. Training was conducted on NVIDIA RTX A6000 GPUs and the Fully Sharded Data Parallel (FSDP) protocol was used to train across GPUs. Across all experiments, we maintained 30 rollouts for GRPO to mimic the 30 mutations generated by the original ES study by \citet{qiuEvolutionStrategiesScale2025}. To benchmark-finetuning, we used a batch-size of 200 examples along with a mini-batch size of 32 examples. A KL-Loss coefficient of $\beta = 0.001$ was used. The trainer was set to run for a total of 500 epochs, although once the validation accuracy appeared to plateau, we stopped training prematurely.
\subsubsection{ES}
We replicated the original author's implementation of ES with two improvements: the authors found that using fp16 instead of bf16 improved validation accuracy on certain tasks. Additionally, the application of the Qwen chat template to the original task prompts improved validation accuracy on the experiment replica Countdown task for Qwen2.5-1.5B, but left model performance on all other regimes virtually the same. Runs were performed both with and without the chat template to assess the effect.


\subsubsection{Reward functions}
For the countdown task, we employ the same reward function used by \citet{qiuEvolutionStrategiesScale2025}, adapted to fit the VERL API. An answer reward is calculated, which assigns a reward of $1.0$ if the model's answer uses all numbers once and evaluates to the provided target, and $0.0$ otherwise. A separate format score is calculated, which serves to ensure that the model's response obeys an XML-style format with \verb|<think>...</think>| thinking tokens first followed by response tokens \verb|<answer>...</answer>|. We take a weighted average of the two rewards to calculate the final reward to assign to the model: $\mathrm{Reward}
= 0.1 \cdot \mathrm{Format\ Reward}
+ 0.9 \cdot \mathrm{Answer\ Reward}
$

For the GSM8K, MATH, and OlympiadBench benchmarks, we employ a rule-based reward function using a binary evaluation logic. An answer reward is calculated by extracting the model's conclusion from the final 300 characters of the response using a regex pattern. The function first identifies the \verb|#### [number]| format, falling back to \verb|\boxed{...}| tags if necessary, and assigns a reward of $1.0$ if the extraction matches the ground truth and $0.0$ otherwise. 

\subsection{Hyperparameter Values}
\begin{table}[H]
\centering
\begin{tabular}{lc}
\toprule
\textbf{Hyperparameter} & \textbf{Value} \\
\midrule
Population size & 30 \\
Noise scale $\sigma$ & 0.001 \\
Learning rate $\alpha$ & 0.0005 \\
Max tokens & 1024 \\
\bottomrule
\end{tabular}
\caption{Hyperparameters used for Evolution Strategies (ES) fine-tuning.}
\label{tab:es-hyperparameters}
\end{table}

\subsection{Additional Experiments} \label{appendix:additional-exp}
\subsubsection{Catastrophic Forgetting and KL} \label{appendix:additional-exp:cat-forgetting-kl}

\begin{figure*}[t]
  \centering
  \includegraphics[width=\textwidth]{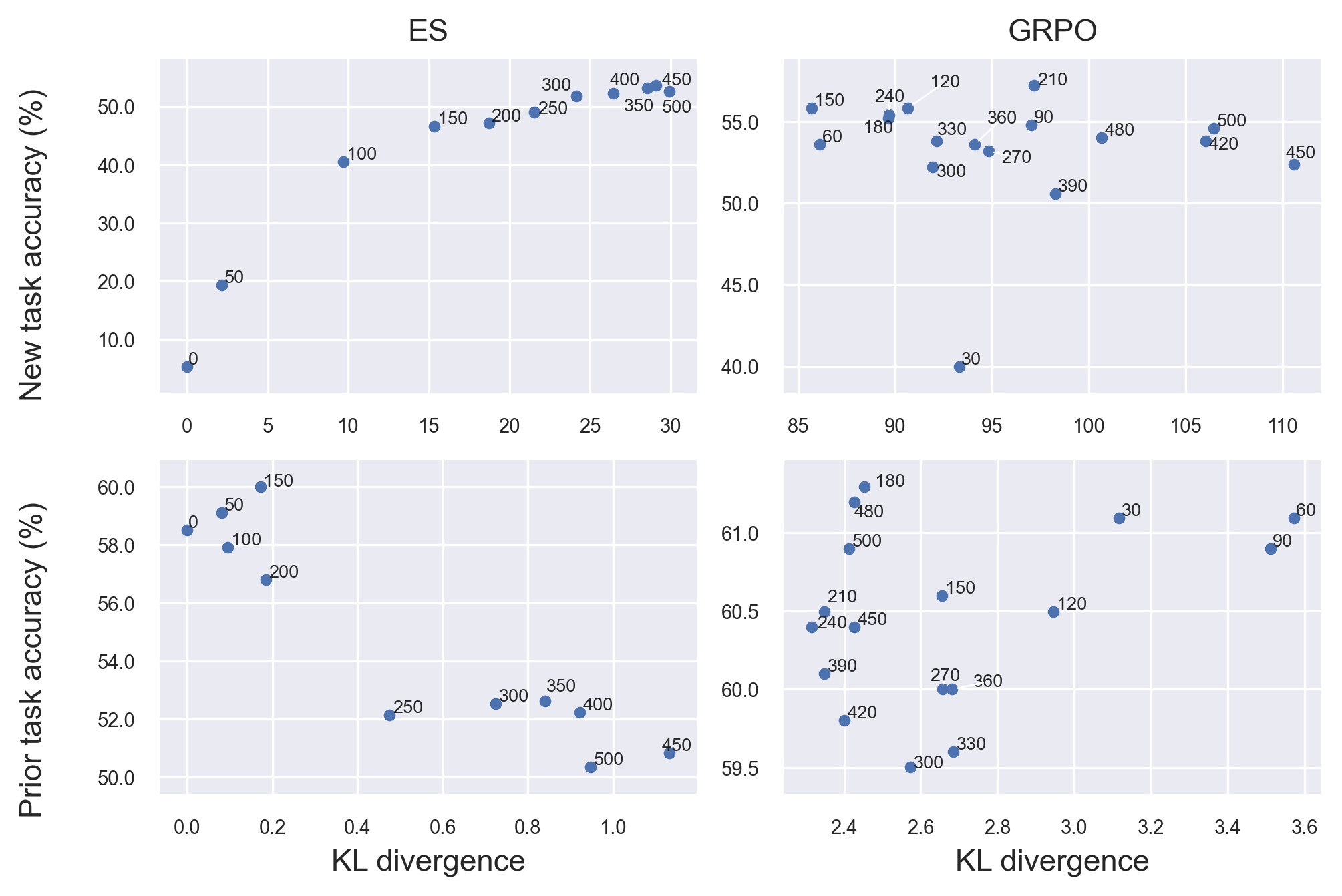}
  \vspace{-0.4em}
  \caption{
  Relationship between KL divergence and task performance.
  Top row: new-task accuracy (Countdown).
  Bottom row: prior-task accuracy (HellaSwag).  Training step indicated per sample.
  ES exhibits increasing KL accompanied by degradation on the prior task, whereas GRPO maintains stable performance across a broader KL range.
  }
  \label{fig:kl-combined}
\end{figure*}

\citet{shenfeld2025rlsrazoronlinereinforcement} had previously established a negative correlation between KL-divergence and previous task score.  We therefore searched whether this trend also is reflected within ES-trained models .  We first looked at KL-divergence between the trained and base models \ref{fig:kl-combined} on the newly trained task.  While ES-trained models increase in KL-divergence with subsequent training steps, this behavior was not consistent when trained with GRPO.  This can be attributed to the explicit KL-regularization factor in GRPO, preventing continuous drifts from the base model.  

The trends for KL divergence and accuracy continue to diverge when evaluating previously known tasks \ref{fig:kl-combined}.  ES has a clear negative correlation between KL-divergence and old task performance.  The KL-divergence between the new and base models also shows an increase over number of training iterations in ES.  GRPO however continues to show no association between number of training steps and KL-divergence, as well as between KL-divergence and previous task accuracy.  Therefore, KL-divergence is a less reliable indicator of catastrophic convergence across GRPO and ES.

\begin{figure}[t]
\centering
\includegraphics[width=\linewidth]{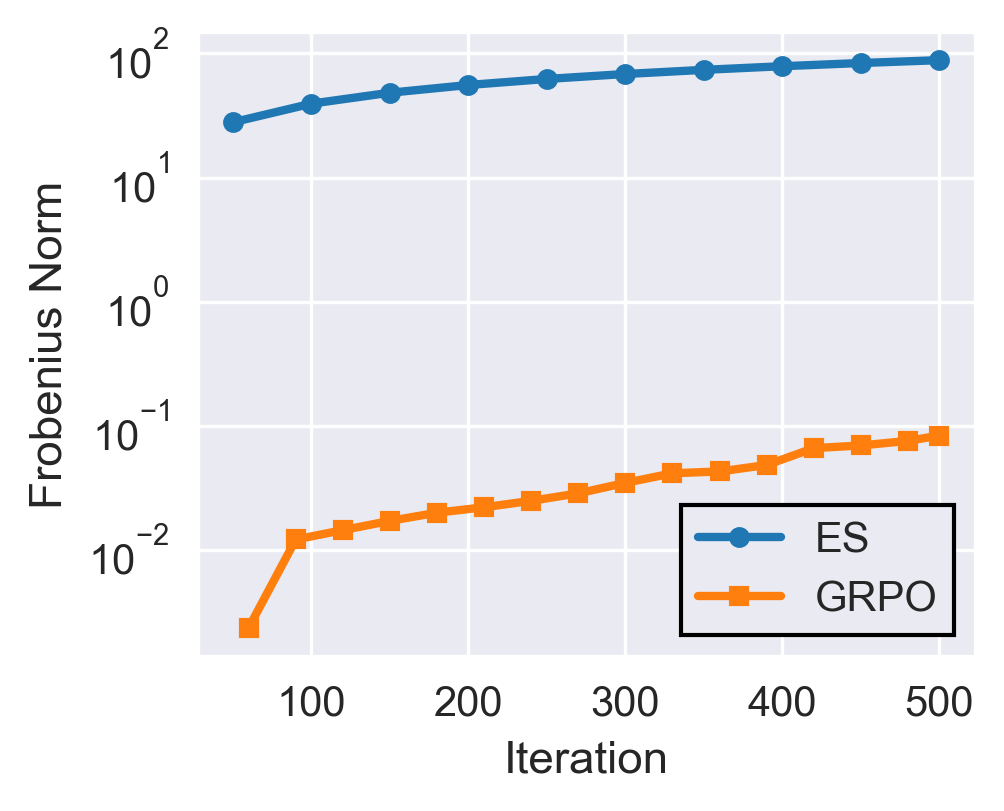}
\caption{
Log relationship between Frobenius norm of a model update and number of training iterations on a new task (Countdown).  ES-trained models drift several orders of magnitude more than GRPO-trained models.  
}
\label{fig:hellaswag-frobenius-log}
\end{figure}

\begin{figure*}[t]
\centering

\vspace{0.4em}
\begin{minipage}[t]{0.49\textwidth}
  \centering
  \includegraphics[width=\linewidth]{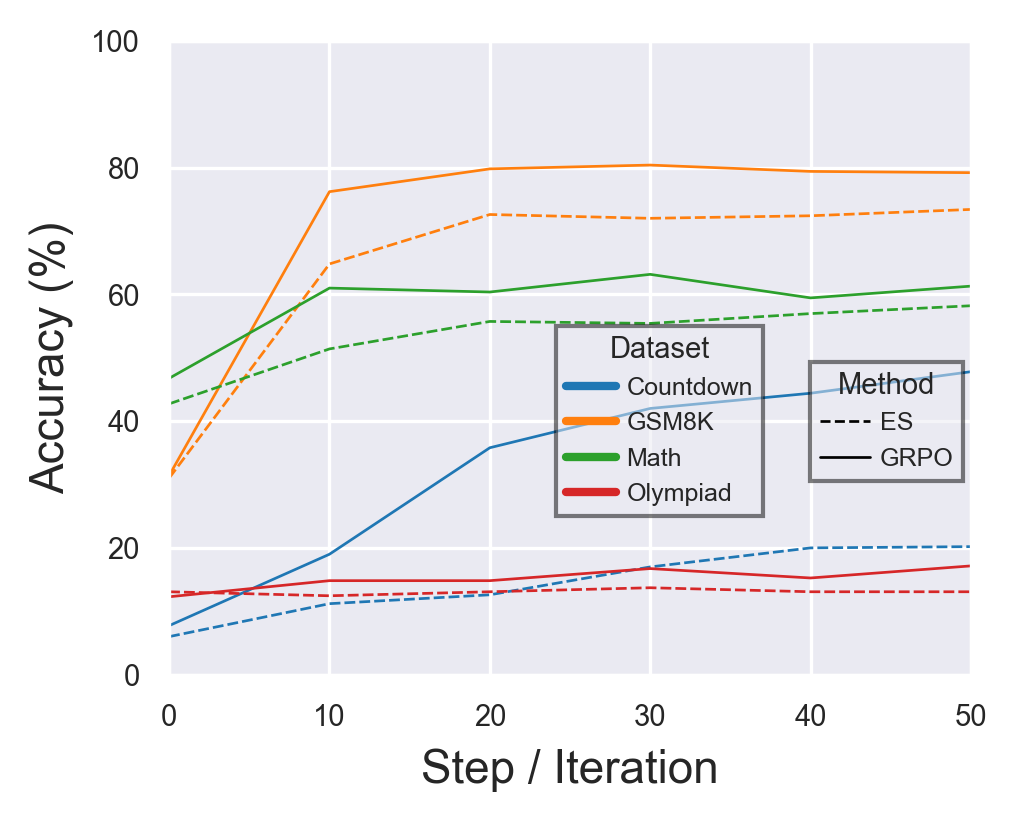}\\[-0.2em]
  {\footnotesize\textbf{(a)} Qwen-2.5-1.5B}
\end{minipage}
\hfill
\begin{minipage}[t]{0.49\textwidth}
  \centering
  \includegraphics[width=\linewidth]{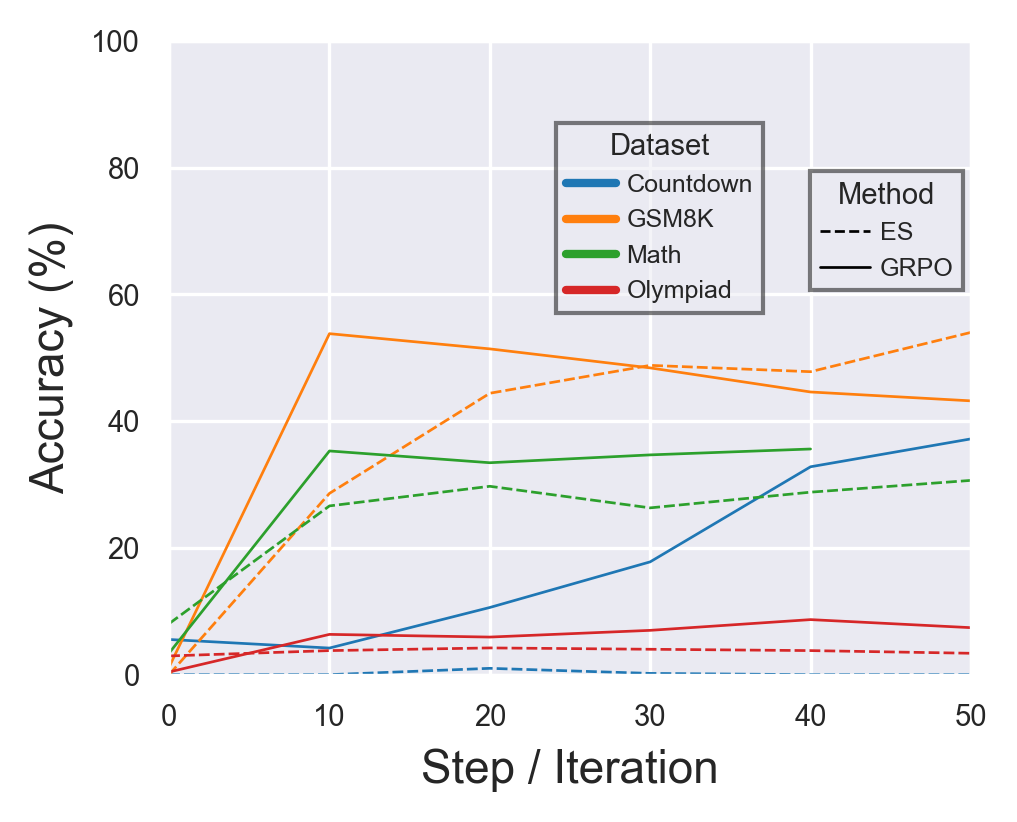}\\[-0.2em]
  {\footnotesize\textbf{(b)} LLaMa-3.2-1B}
\end{minipage}
\vspace{-0.6em}
\caption{
Mean accuracy curves for ES and GRPO runs across across datasets: Countdown, GSM8K, Math, Olympiad.
}
\label{fig:es-grpo-comparison}
\end{figure*}




\end{document}